\documentclass{article}

\usepackage{bm}
\usepackage{amssymb}
\usepackage{graphicx}

\begin{document}

\title{Discovering indicators of dark horse of soccer games by deep learning from sequential trading data}

\author{Liyao Lu \\
Walmart Labs \\
Sunnyvale, CA, USA \\
liyaolu3@gmail.com \\
\and
Qiang Lyu \\
School of computer science and technology, Soochow University \\
Suzhou, Jiangsu, China \\
qiang@suda.edu.cn}

\date{}
\maketitle

\begin{abstract}
It is not surprise for machine learning models to provide decent prediction accuracy of soccer games outcomes based on various objective metrics.
However, the performance is not that decent in terms of predicting difficult and valuable matches.
A deep learning model is designed and trained on a real sequential trading data from the real prediction market,
with the assumption that such trading data contain critical latent information to determine the game outcomes.
A new loss function is proposed which biases the selection toward matches with high investment return to train our model.
Full investigation of 4669 top soccer league matches showed that our model traded off prediction accuracy for high value return due to a certain ability to detect dark horses.
A further try is conducted to depict some indicators discovered by our model for describing key features of big dark horses and regular hot horses.
\end{abstract}


\section{Introduction}
Sports analytics has been well studied and applied in various kinds of sports games since the 1970s \cite{Stefani1977}.
The analytic technologies are evolving from statistical to computational approaches \cite{MLSA2017}.
Applying machine learning to soccer games analytics has brought more and more attention to both the sports industry and computing academia.

Traditionally, people assess soccer players' and teams' attributes via quantified ratings by experts, big sports media, and professional league web sites based mainly on statistical methods.
For example, Sky Sports, the English Premier League broadcast mogul, developed the \emph{Power Ranking} system to calculate Premier League players' overall performance based on 32 statistical features \cite{skysports2017}.
Such objective metrics \cite{Hughes2007} are usually main data sources for people making decisions associated with soccer games.
This classical scenario is vividly exhibited in the most famous management simulation computer game \emph{FootBallManager} \cite{footballmanager}.

Predicting the outcome of a soccer match is usually a main task for both academia and industry \cite{Haaren2011}.
For matches covering 2014 World Cup finals, 2012 UEFA European Championships and 2015 Copa America,
based on Elo \cite{Elobook},  FIFA Women¡¯s World Ranking Methodology \cite{FIFAranking} and similar ratings, regression models got prediction accuracy ranging from 50\% to 55\% \cite{Lasek2016}.
Besides academia, Tech giants such as Microsoft Bing  also participated in predicting outcomes of English Premier League every season \cite{bingpred}.
Bing correctly predicted 125 matches out of 232 in 2017, corresponding to an accuracy of 53.88\%.
It is marketed as a satisfying performance in terms of game outcome prediction accuracy.

Such predictions comfortably outperformed the random guess accuracy of 1/3 for three-way soccer games.
However, accuracy alone could be misleading to evaluate the prediction quality and performance, because by doing so, it is implicitly assumed that the difficulty and value of each correct prediction are equal, which is actually untrue.
It is much easier to correctly predict a match's outcome when one team is much stronger than its opponent, such as the \emph{La Liga} match Barcelona VS Celta Vigo,
than a match when two teams are close, such as the \emph{EL CLASSIC} match Real Madrid VS Barcelona.
Correctly predicting a Barcelona victory in the first match will receive less award from prediction market than correctly predicting the \emph{EL CLASSIC} match.
Basically, the rate of return based on the prediction results can be thought of as quantitative indicator of the difficulty and value of each correct prediction.
Optimizing reward is much more difficult than optimizing prediction accuracy.
For example, if we place \$1 in Betfair's prediction market \cite{betfair} on each of the 232 Premier League matches in 2017 guided by Microsoft's prediction,
the return gain is \$228.23, which is only 98.37\% of our total stake.

The reason for previous unprofitable predictions against the money line is probably not algorithmic,
but because of lacking insider information that indicates players' actual status of the coming match \cite{Levitt2004}.
Insider information such as morale, locker room stories, and unannounced injuries affect game outcomes significantly but are unknown to the public \cite{Croxson2014}.
In contrast, it is not hard for bookmakers and big market makers to obtain up-to-date insider information since they sponsor many leagues and teams.
Based on their predictions with up-to-date insider information, they maximize profits and minimize potential loss by actions including bidding, trading, and modifying odds in prediction markets \cite{SMITH2009539}.

It is believed that the market is not manipulated by individual participants, and bids information enriches some latent factors to determine the outcomes of some specific matches \cite{Dixon1997,Franck2011}.
So in this paper, we used the sequential trading data, instead of performance metrics from teams' side, as our analytic target data \cite{leitner2011bookmaker, STRUMBELJ2014934}.
Therefore, we could probably offset our disadvantage of lacking insider information by digging out the hidden message behind public bids data in prediction market.
We developed a deep learning model to maximize the valuable predictions.
In order to do that, we defined a new objective function, or loss function for sake of training algorithm, which biases the selection toward matches with more return.
Thus our model had a certain ability to detect dark horses.

Generally speaking, dark horse refers to the event with small probability to occur.
In real life, correctly detecting dark horse usually results in significant effects.
In our context, dark horses refer to those matches whose outcomes are less likely to happen and thus returns are higher than other matches.
Our loss function and learning model tried to capture those dark horses.
We fully investigated the average learning ability of our model on a real sequential trading data set, containing 4669 soccer matches of top soccer leagues.
Results showed that our model underperformed in terms of outcome accuracy but outperformed in terms of valuable return,
which reflects our model's ability to detect dark horses.

In a pilot study, similar learning approach was applied to sequential bids data from Bookmaker \cite{Lyu2018}.
We believe that trading data should contain more information than bids data.
Bids data are static intends of bettors' expression.
Although in context of time they can form a dynamical sequential data,
bids data lack important deal volume information.
For example, a high bid without any deals might leak different information from a bid with high deal volumes.
It seemed to suggest that even with the same learning model, learning from trading data flow might be easier than learning from bids data flow,
since trading data contained deals information which bids data did not.
Thus we expect that more useful information will be exploited by learning from sequential trading data.

Overall, our research makes three contributions.
First, we targeted sequential sequential trading data which are believed to embed rich pattern and latent information.
Second, we aimed at valuable predictions instead of regular accuracy.
And as a natural consequence, we achieved some ability to identify dark horses.
Last but not least, we tried to depict some key indicators discovered by our model for describing key features of big dark horses and hot horses.

The rest of paper is organized as follows:
after the problem formulation and data set description are given in Section 2,
a deep learning model and new loss function are presented in Section 3.
Results and evaluations are fully described in Section 4.
And further discussion ends the paper.

\section{Problem formulation and data set}
Let $y \in \{draw, win, lose\}$ be the final outcome of a soccer match in terms of home team against guest team.
For each soccer match, there are three $Odds_y$s corresponding to three possible outcomes respectively at given time $i$.
If participants bet \$1 on $y$, then they will get $Odds_y$ back if $y$ eventually occurs as the match's outcome. Otherwise, they will lose the bet.
Given a bids data sequence $x_1, \ldots, x_T$ for each match, where $x_i$ contains various bids data at time $i$ and $T$ is the final second before the match, our goal is NOT to predict the outcome $y$ of the match.
Instead, our final goal is to maximize the expectation of gains $G$ for betting a group of matches.
That is,
given the bids data sequences of a group of $N$ matches, for each match $m$, at the time $T$ with three $Odds$ of the match outcomes, we make a \$1 bet on one prediction $\hat y^m$.
Our final goal is to let
\begin{equation}
\label{eq:obj}
G = \max{\frac 1 N \sum_{m=1}^N \left ( Odds_y^m * \delta(y^m,\hat y^m) \right )},
\end{equation}
where $\delta$ is the identity function and $y^m$ is the real final outcome of match $m$.

It is important to point out that our final goal described in equation (\ref{eq:obj}) is partially different from the conventional goal of a machine learning task.
Usually, a machine learning task aims to maximize the outcome prediction accuracy of the matches.
Although $G$ in equation (\ref{eq:obj}) is conceptually in direct proportion to the prediction accuracy,
our model prefers correctly betting on one match with higher $Odds^m$ to two matches with lower $Odds^j$ and $Odds^k$ respectively, when $Odds^m > Odds^j + Odds^k$.
In this paper,
we refer the match whose eventual outcome agrees with the biggest $Odds$ to a \emph{big dark horse}.
While the match whose eventual outcome agrees with the smallest $Odds$ is called \emph{hot horse},
and \emph{middle horse} is the match whose eventual outcome agrees with the $Odds$ stands between the biggest and smallest $Odds$.
We generally refer \emph{dark horse} to the union of \emph{big dark horse} and \emph{middle horse}.

By modeling the primary goal as equation (\ref{eq:obj}),
we intentionally try to maximizing the gain,
and unintentionally make the prediction accuracy as the secondary goal.
That is to say, prediction accuracy becomes a mean instead of an ultimate goal in this paper.
This is why we call our solution of maximizing gain as an end-to-end solution.

We bought real bids data of Betfair \cite{betfair} from its licsensed data agent company Fracsoft \cite{fracsoft}.
Betfair is a prediction market platform for client-to-client trading, similar with stock exchange platform.
Every participant can bid buy/sell prices and volumes on that platform, and of course trade any available bids.
Our data set contains bids data sequence of English Premier League from 2007 to 2014,
Spanish \emph{La Liga} League from 2008, 2010 to 2014,
and France \emph{Ligue 1} League from 2011 to 2014.
These leagues are all top soccer leagues of their own countries and of the world.
The missing data of some years were due to the data provider's business restriction.

We collected some trading data for each time interval from the raw data.
There are mainly three feature vectors within an $x_i$, describing the trading information between time $i-1$ and $i$.
These vectors describe trading data at time $i$ for a match outcome: $win, lose$ and $draw$ respectively.
Each feature vector summarizes trading information of two basic groups, (Back, Buy) and (Lay, Sell), occurred within a certain time interval in a prediction market.
People who short a certain outcome can submit Back bids with some volumes at a certain odds, so that someone else who long that outcome can Buy.
In contrast, people who long a certain outcome can submit Lay bids and someone else who short that outcome can Sell.
Both Back and Lay bids can be cancelled before they are matched by buyers or sellers.
We use 4 features to summarize all Buy actions occurred in a certain time interval according to Table \ref{tab:buyfeatures}.

\begin{table}[hptb]
\centering
\caption{Features related to Buy with description}
\label{tab:buyfeatures}
\begin{tabular}{ll}
\hline
Feature      & Description                                      \\
\hline
BuyActionCnt & The number of times of Buy                   \\
BuyVolAvg    & The mean volume of Buy                  \\
BuyVolStd    & The standard deviation of Buy volume  \\
BuyOddsAvg   & The average odds of Buy  \\
\hline
\end{tabular}
\end{table}

In addition, 6 features are applied to summarize all Back actions occurred in a certain time interval according to Table \ref{tab:backfeatures}.

\begin{table}[hptb]
\centering
\caption{Features related to Back with description}
\label{tab:backfeatures}
\begin{tabular}{ll}
\hline
Feature    & Description                                 \\
\hline
BackBidsSubmitted   & The number of Back bids                    \\
BackSubmittedVolAvg & The mean volume of Back bids               \\
BackSubmittedVolStd & The std volume of Back bids \\
BackBidsCancelled   & The number of cancelled Back bids                     \\
BackCancelledVolAvg & The mean vol. of cancelled Back bids              \\
BackCancelledVolStd & The std vol. of cancelled Back bids \\
\hline
\end{tabular}
\end{table}

So there are 10 features in a group (Back, Buy).
Similarly, there are another 10 features in a group (Lay, Sell) counterpart.
In summary, there are 20 features for a feature vector of a match outcome,
and the total dimension of $x_i$ is $20*3=60$.

It is noted that deal volume information and bids cancelling information are all included in the feature vector.
These are the features totally different from bids features.

Since the data sequence varied in length and trading frequency,
we need to preprocess the raw data.
Firstly, for a match we truncated trading data sequence to keep all valid data of 2 hours before the opening whistle.
Secondly, we dropped matches which had too few trading data.
Lastly, we sampled the sequential trading data with the sampling strategy described in Table \ref{tab:freq}.
\begin{table}[hptb]
\centering
\caption{Sample strategy to generate sequential trading data}
\label{tab:freq}
\begin{tabular}{lll}
\hline
sample period  & time interval     & sample points                                 \\
\hline
1st             & 10 seconds        & 90 before the match begins \\
2nd             & 20 seconds        & 90 before the 1st period \\
3rd             & 30 seconds        & 59 before the 2nd period \\
4th             & till available     & 1 before the 3rd period \\
\hline
\end{tabular}
\end{table}

Consequently for all matches, the length of a sequence of trading data is regulated to 240.

After filtering out some matches with error data, we have a clean data set of size 4669 matches for the rest of this paper.

\section{Designing and training the deep model}
The learning model is designed as the following expression:
\begin{equation}
\label{eq:model}
y = \textnormal{MLP} \left ( \textnormal{CONCAT}(\textnormal{RNN}(\textnormal{CNN}(\bm{x_t})),\bm{x_s}) \right ),
\end{equation}
where $\bm{x_t}$ is sequential trade data flow of a match, $\bm{x_s}$ is non-sequential features of the match,
CNN stands for a block based on convolutional neural network \cite{LeNet5Paper},
RNN for a block based on recurrent neural network \cite{lstmPaper},
CONCAT for concatenation and MLP for a block based on multiple layer perception \cite{MLPOriginalPaper}.
Basically, the raw data flow is first feed to CNN for mining new features based on neighbors.
So we call such features local features.
Then the sequentially mined local features are forward to RNN for accumulatively extracting features as the representative features of the whole sequence.
We call features from RNN global features.
Finally the global features and other non-sequential features $\bm{x_s}$ are combined and input to MLP for constructing a classifier.
$\bm{x_s}$ is different from the local and global features in that it includes all state features for the match.
The order of features in $\bm{x_s}$ does not matter for the learning task.
Such example features of $\bm {x_s}$ are League type (to which country does this league belong) and match type (strong team against strong, or strong against week team, and so on).

CNN consists of several 1D convolutional layers.
The first layer is defined as
\begin{equation}
\label{eq:cnn1}
\bm f^1 = \textnormal{sigmoid}[\mathcal C^1_{9}(\bm{x_t})],
\end{equation}
where the subscript of the convolution operand $\mathcal C$ stands for the number of operands and superscript for the window size.
The purpose of the first layer is to reorganize the raw features with a non linear activation function \cite{NiNPaper}.

The second layer of CNN is defined as
\begin{equation}
\label{eq:locallyC}
\bm f^2 = \mathcal P_{max}^2 \left ( \textnormal{sigmoid}[\mathbb C^5_{3}(\bm f^1)] \right ),
\end{equation}
where the special convolution operand $\mathbb C$ here is totally different from the regular convolution operand $\mathcal C$ in
that $\mathbb C$ does not share weights along each time spot of the sequence like a regular $\mathcal C$ does.
This means that for different time spots the extracting rules are allowed to be different.
It is very likely that the bids data close to the match beginning time are differently embedded with feature from that far from the match beginning time.
So this layer is expected to extract more useful features.
The operand $\mathcal P_{max}^2$ in equation (\ref{eq:locallyC}) denotes a max pooling operation with pooling size 2.
This makes the length of the sequence $\bm f^1$ shortened in half to $\bm f^2$.

The third layer of CNN is similar with the first layer by defining
\[
\bm f^3 = \mathcal P_{max}^2 \left ( \textnormal{reLu}[\mathcal C^3_{3}(\bm f^2)] \right ),
\]
where reLu is rectified linear unit \cite{relu} function used as the activation output of the third layer.

The three layers of CNN explore completely different dimensions for extracting local features.
The first is focused on the $x_i$ internal.
The second aims at different time spot.
The last is trying to capture features along the time axis.

RNN block is defined as
\[
\bm f^4 = \textnormal{reLu} [ \textnormal {GRU}_{9}(\bm f^3) ],
\]
where GRU is a gated recurrent network \cite{gruPaper},
and the subscript 9 denotes the output dimension.
GRU is a simplified implementation of LSTM \cite{lstmPaper}.

CONCAT($\bm f^4, \bm {x_s}$) just concatenates the two input vectors.
$\bm {x_s}$ obviously needs manually annotation on the raw data, therefore it embeds human's subjective intelligence.
In this study, since we are focusing on learning from the objective raw data, we let $\bm {x_s}$ be null.

MLP block consists of three fully connected regular neural networks, as defined by
\begin{equation}
\label{eq:mlp}
y = \textnormal{softmax} \left (\mathcal D_3 (\mathcal D_{18} (\bm f^4))) \right ),
\end{equation}
where $\mathcal D$ is a layer consists of a fully connected neurons whose number is represented as the subscript.

In order to train a regular machine learning model like equation (\ref{eq:model}),
we define our own loss function for the learning model (\ref{eq:model}) as the following:
\begin{equation}
\label{eq:loss}
loss =  \frac 1 B \sum_{m=1}^B \left ( Odds_y^m * \mathcal E(\hat y^m) \right ) + \lambda_1 \mathcal L_1(\theta) + \lambda_2 \mathcal L_2(\theta),
\end{equation}
where $B$ is mini batch size, $\mathcal E(\hat y^m)$ is the entropy of predicting probability $\hat y^m$,
and $\theta$ are the parameters of our learning model.
$\mathcal L_1$ and $\mathcal L_2$ are norm 1 and norm 2 respectively,
and $\lambda_1$ and $\lambda_2$ are corresponding weights of $\mathcal L_1$ and $\mathcal L_2$.

Please note the subtle difference between the regular categorical cross entropy loss and our loss defined in equation (\ref{eq:loss}).
The regular cross entropy loss function gives each label the same importance,
while our loss gives dark horse more weights.
This is the root why our model might have more chance to catch dark horse than the model using regular cross entropy loss function.

By now, any backpropagation based training algorithms \cite{Nature1986Paper} can be used to minimize our loss function for learning model (\ref{eq:model}).
For convergence checking, we monitor the decrease of validation loss.
If the validation loss does not decrease for a continuous 8 epochs, we consider the training process is convergent and then stop the training.
We then use the model parameters of when the validation loss is the minimum as our final trained model for the evaluations.

\section{Results}
For a not-so-large data set, tuning the best performance model on a fixed test set is usually possible.
However, this does not ensure the good generalization of the trained model on the data outside the test set.
In this study, we focused on evaluating the average learning ability of our model by the following design.
We run multiple independent trials on the whole data set.
For each trial, we randomly selected 10\% of the data set as the test set and the rest as the training set.
This makes the test set size of 467 matches in this section.
Among the training set, a random 10\% was selected as the validation set.
Based on all these trials, we tried to analyze the average performance for evaluating our model in this section.
This evaluation strategy is a variant of cross validation strategy, but with more fine granularity.

We used Keras \cite{KerasURL} as our front programming framework, and Tensorflow \cite{abadi2016tensorflow} as the underlined deep learning engine.
For the other training parameters setting,
we set $B=64$ as the mini batch size in equation (\ref{eq:loss}),
and $\lambda_1 = \lambda_2$=1e-3.
We used Adam optimization algorithm \cite{AdamPaper} to minimize equation (\ref{eq:loss}) with learning rate=1e-4 and decay=1e-5.

We run our program on a server equipped with two Nvidia GPU cards, Tesla K20C.
The server has two 12-core E5-2620 CPUs with 64GB memory.
It took 11 seconds for an epoch learning for the above settings.

\subsection{Baselines setting and bet policies}
For results evaluation in terms of valuable predictions, we set up five baselines.
The first is the gain $G_{rand}$ based on random guess strategy, which is the gain of betting on the random selection of $win, lose$ or $draw$.
The second is the gain $G_{min}$ based on min-Odds guess strategy, which is the gain of betting on the outcome with minimum odds.
According to \cite{bland2000odds}, the probability of a certain outcome is roughly the reciprocal of its odds.
Thus, $G_{min}$ is considered as a naive strategy that picks the outcome seemingly most likely to happen according to the static pre-game odds.
Similarly, the third and fourth baselines are $G_{max}$ and $G_{middle}$.
It seems that $G_{min}$ and $G_{rand}$ are rational choices if no other information are available for making decision.
The last is the best gain $G_{best}$, which is the gain of betting all correctly with the outcome of the match.
$G_{best}$ is of course the ceiling line which is never touched by any predictions.

For each test set we used three policies based on the prediction probabilities to evaluate the gains of our trained models.
One-bet policy (1-bet for short) means that we bet \$1 on one of the three outcomes, $win, lose$ or $draw$, by selecting the max probability of the prediction.
Split-bet policy (s-bet for short) means that we split \$1 on three bets according to the three prediction probabilities of three outcomes.
In fact, s-bet can be thought as of a hedge policy.
It will not get nothing or highest return no matter what outcome occurs.
It is very easy to see that our loss function defined in equation (\ref{eq:loss}) is in favor of 1-bet.
Including s-bet results here is just for evaluation.
The third bet policy is called dark horse policy (d-bet for short),
in which case we ignore those matches whose predictions that our model agreed with $G_{min}$,
then we apply 1-bet to the rest of the matches.

\subsection{Computational results}
We do the following evaluations based on 87 random trials, and ensure that every match has chance to be in the test set at least one time.

\begin{figure}[hptb]
  \centering
  \includegraphics[scale=0.5]{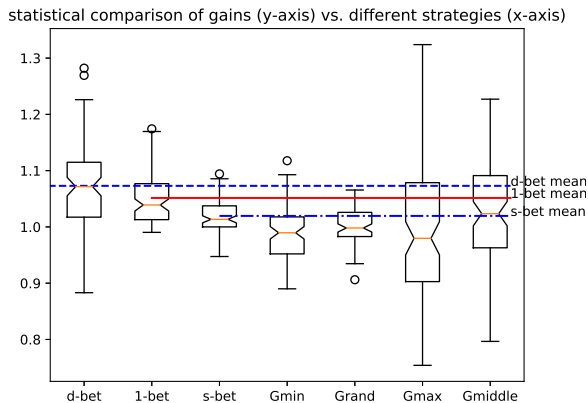}\\
  \caption{Overall performance comparing with the baselines}
  \label{fig:box}
\end{figure}

It is easy to accept that the ideal expectation of $G_{rand}$ is close to 1.
The interesting thing is that $G_{min}, G_{max}$ and $G_{middle}$ are all close to 1, and $G_{best}$ is close to 3 \cite{Cortis2015}.
Check Figure \ref{fig:box} for the empirical results.

Since $\frac 1 {Odds}$ can be interpreted as the probability of a outcome \cite{bland2000odds},
and for the ideal fair condition,
\[
P_{max}+P_{min}+P_{middle} = 1,
\]
where $P_{max}=\frac 1 {Odds_{min}}$ is the outcome choice of $G_{min}$.
The accuracy of $G_{min}$ depends on the ratio of dark horses over all matches.
It is nature that the probability of dark horses is less than that of hot horses.
So the accuracy of $G_{min}$ is always a little above 50\%.

Figure \ref{fig:box} showed the overall performance comparison between our model and baselines.

The first three boxes in Figure \ref{fig:box} showed our model's better performance over the baselines.
The average gains of d-bet and 1-bet were close to 1.07.
We were very pleased that the average gains of our model were superior to the first quartile of $G_{min}$ and $G_{rand}$.
1-bet obtained the solid and good performance among all the gains because d-bet took risk for extreme returns.
The following investigation was now based on 1-bet performance.

\begin{figure}[hptb]
  \centering
  \includegraphics[scale=0.5]{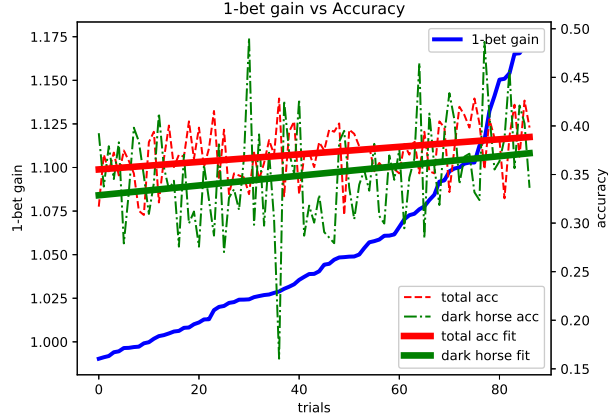}\\
  \caption{Sorted 1-bet gains vs prediction accuracies}
  \label{fig:gacc}
\end{figure}

An interesting point was such gains were obtained under the condition of prediction accuracies less than 50\%.
Figure \ref{fig:gacc} showed the relationship between gain and prediction accuracies.

The average of total prediction accuracies was about 39\%, and that of dark horse accuracy was about 34\%.
The slope of dark horse accuracy was a little bit sharp than that of total prediction accuracy,
which demonstrated that dark horse accuracy contributed more to gain than the total prediction accuracy.

\begin{figure}[hptb]
  \centering
  \includegraphics[scale=0.5]{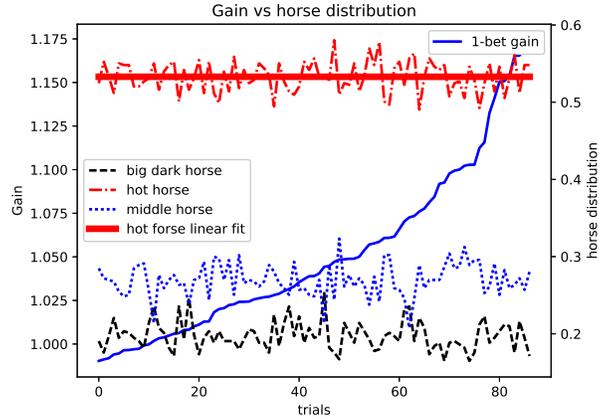}\\
  \caption{Sorted 1-bet gains vs horse distribution}
  \label{fig:ghDist}
\end{figure}

Considering the fact that $G_{rand}$ is almost equal to $G_{min}$ despite the prediction accuracy varying from 33\% to 53\%, shown in Figure \ref{fig:ghDist},
it is understandable that our model had chance to get better gains with the accuracy less than 50\%
because we captured more dark horses than $G_{rand}$ and dropped more hot horses than $G_{min}$.

\begin{figure}[hptb]
  \centering
  \includegraphics[scale=0.5]{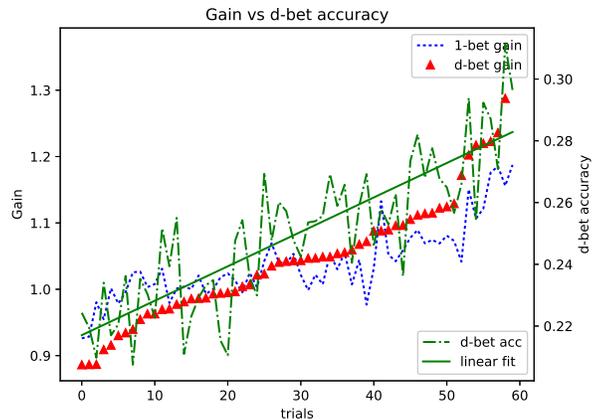}\\
  \caption{Sorted d-bet gains vs prediction accuracy}
  \label{fig:dgacc}
\end{figure}

The importance of detecting dark horse correctly was illustrated in Figure \ref{fig:dgacc},
where we adopted the more aggressive bet policy, d-bet.
Figure \ref{fig:dgacc} showed that d-bet gains was better than 1-bet even with the average prediction accuracy as low as 26\%.
But it was acceptable that d-bet was not so stable than 1-bet by observing Box 1 and 2 in Figure \ref{fig:box}.

We further wanted to know why our model had different accuracies of detecting dark horses.
Figure \ref{fig:ghDist} showed how 1-bet gains related to the horse distribution.

First, Figure \ref{fig:ghDist} told that the three distributions of three types of horses in each test set were almost uniform.
This ensured us the uniform partition between the train set and test set.
Second, it was easy to know that the three distributions were also the prediction accuracies corresponding to the three baselines,
hot horse distribution for predicting accuracy of $G_{min}$,
big dark horse distribution for $G_{max}$,
and middle horse distribution for $G_{middle}$.
Of course the prediction accuracy of $G_{rand}$ was 33\%.
We wanted to emphasize that despite the large difference of these accuracies, their average gains were not so much different.
It also proved that predicting hot horse correctly was easier but less valuable than predicting dark horse correctly.
Third, the linear fit of hot horse distribution in Figure \ref{fig:ghDist} showed a little bit decrease,
which reflected the overall increase of dark horse distribution.
So this explained the reason of the increase trend of dark horse accuracy in Figure \ref{fig:gacc}.
It also told that the ability of our model to detect dark horse was not random but stable.

\begin{figure}[hptb]
  \centering
  \includegraphics[scale=0.5]{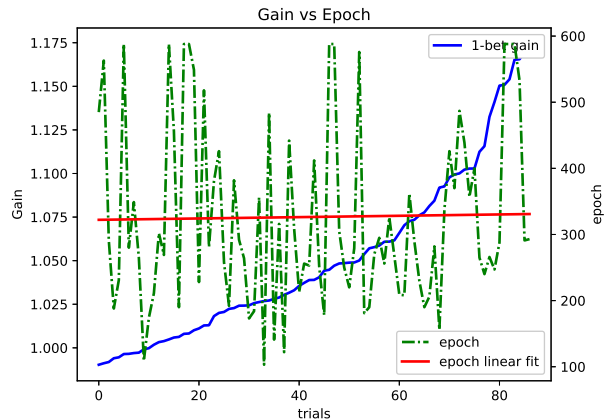}\\
  \caption{Sorted 1-bet gains vs convergent epochs}
  \label{fig:gepoch}
\end{figure}

Finally, we wanted to investigate whether the higher 1-bet gains benefited from longer training.
Figure \ref{fig:gepoch} showed convergent epochs of all the experiment trials.

The linear fit of epochs in Figure \ref{fig:gepoch} was independent with the increase of 1-bet gains.
This proved that our training was basically stable, as well as the model's learning ability.

\subsection{Discovering indicators}
In this section, we tried to exploit possible indicators of dark horse discovered by our learning model.
The best indicator must be simple and suited for indicating as much dark horses as possible.
Unfortunately, the accuracy for detecting dark horse of our model was about 34\%.
It seems not practical to discover the ideal consistent indicators for all dark horses.
We turned to try finding the possible differentiable indicators between the darkest and the hottest horse.
The following analysis was based on a typical trial.

\begin{figure}[hptb]
\centering
\includegraphics[scale=0.5]{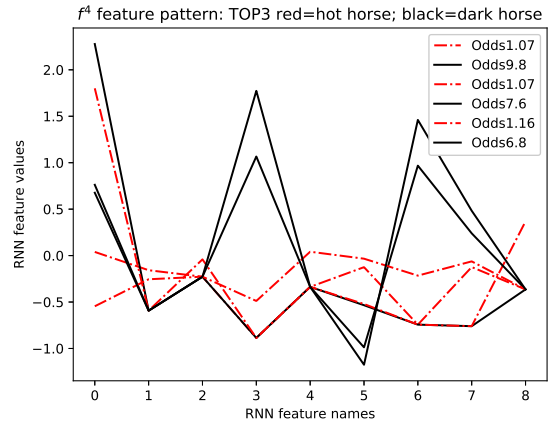} \\
\caption{Feature patterns learned from RNN block for top 3 hottest and darkest horses.}
\label{fig:rnn}
\end{figure}

We started from the learned features from CNN and RNN blocks,
which automatically extracted 9 features to MLP for constructing a classifier according to equation (\ref{eq:mlp}).
First, we analysed the 9 distributions of these features' values.
We failed to seek some special distributions of specific feature between dark horses and hot horses.
However, when we depicted the pattern of these 9 features of TOP 3 darkest and hottest horses in Figure \ref{fig:rnn},
the differentiable pattern was revealed.

The pattern jointly determined by RNN feature 0, 3, 5, 6 and 7 was clearly different between dark and hot horses in Figure \ref{fig:rnn}.
The darkest horse was a match of Premier League played on February 09, 2014.
The home team was Manchester United football club, and the guest was Fulham football club.
The final outcome of this match was 2-2, a surprising draw.
The procedure of this match was as the following.
The first goal happened at 19 minute of the first half scored by Fulham's Steve Sidwell.
The second and third goals occurred at 78 and 80 minute of the second half from Manchester's Robin Van Persie and Michael Carrick.
At the last minute of match, Darren Bent from Fulham scored the equaliser.

Our model's prediction and pre-game odds were summarized in Table \ref{tab:pred}.
\begin{table}[hptb]
  \caption{Comparison between $G_{min}$ and 1-bet predictions on game Man. Utd vs Fulham}
  \label{tab:pred}
  \centering
  \begin{tabular}{ccccc}
  \hline
            & $win$             & $draw$            & $lose$        & Gain        \\
  \hline
  $Odds_T$  & 1.17              & 9.8               & 22.0          & null       \\
  $G_{min}$ & \textbf{0.8547}$^\dag$ & 0.1020$^\dag$     & 0.0455$^\dag$ & 0     \\
  1-bet     & 0.4075$^\dag$ & \textbf{0.5777}$^\dag$ & 0.0147$^\dag$ & \textbf{9.8}      \\
  \hline
  \end{tabular}

$^\dag$ prediction probability on each outcome.
\end{table}
Our model gave the probability 0.5777 of the outcome $draw$, the highest among the three predictions but a little higher than that of the outcome $win$.
We tried to find how our model made this decision.

\begin{figure}[hptb]
\centering
\includegraphics[scale=0.5]{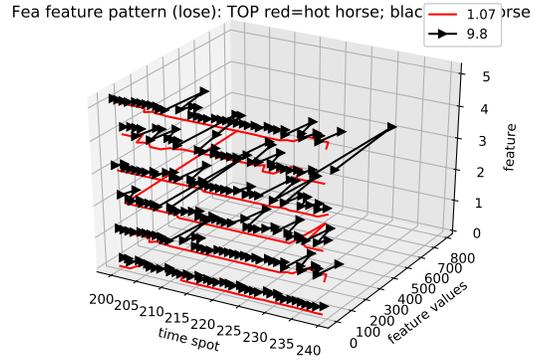}
\includegraphics[scale=0.5]{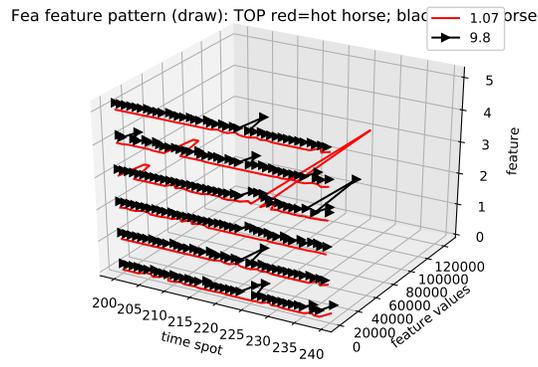}
\includegraphics[scale=0.5]{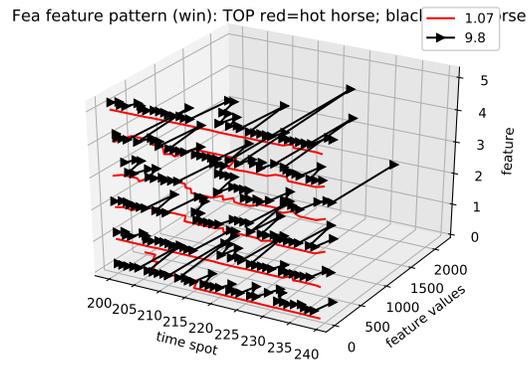}
\caption{Some typical normalized input features.
The top subfigure is for $lose$ outcome, the middle for $draw$ and the bottom for $win$.}
\label{fig:fea}
\end{figure}

We began with visualizing the input feature of the match.
We have 20 features at each timestamp for each of the three possible outcomes, which were previously illustrated in Table \ref{tab:buyfeatures} and Table \ref{tab:backfeatures}.
When BuyVolAvg increases, it indicates that the market feels the outcome more likely to occur eventually.
In contrast with Buy group related features, the trends of the Sell group related curves have the exact opposite meaning.
When BackSubmitted related curves rise, it indicates that the market feels the outcome less likely to occur,
and BackCancelled related curves indicate the opposite market morale.
In addition, the trends of Lay group related curves have the exact opposite meaning with their Back counterparts.
Here we visualized 6 typical features for each outcome in Figure \ref{fig:fea}.

For the clarification reason, we only truncated curves of last 400 seconds.
The map of feature names to z-axis of Figure \ref{fig:fea} is described in Table \ref{tab:visFea}.
\begin{table}[hptb]
\centering
\caption{Feature names related to z-axis in Figure \ref{fig:fea}}
\label{tab:visFea}
\begin{tabular}{lll}
\hline
Feature name                    & z-axis    & favorability of market        \\
\hline
BuyVolAvg                      & 0                    & more \\
SellVolAvg                      & 1                     & less \\
BackSubmittedVolAvg      & 2                    & less \\
SellSubmittedVolAvg       & 3                     & more \\
BackCancelledVolAvg     & 4                     & more \\
SellCancelledVolAvg     & 5                     & less \\
\hline
\end{tabular}
\end{table}

The last column of Table \ref{tab:visFea} indicates the favorability of market if the corresponding feature value is going up.
It was clearly demonstrated in Figure \ref{fig:fea} that the hottest horse had totally different input pattern from the darkest horse.
It is understandable that curves of the hottest horse usually does not change much as the darkest horse.
Besides the frequent changes, there are always inconsistent trends for those dark horses.
That is why dark horse is more difficult to identify than hot horse.
For example, let's check the features curves in Figure \ref{fig:fea}.
Curves of Feature 0 and 1 for $lose$ showed the market was unfavor for this outcome, which was consistent with the market's favor of the other two outcomes.
But checking  these two curves for $win$ and $draw$, both Feature 0 and 1 for $win$ were increasing simultaneously, which was a conflict.
While these two curves for $draw$ were consistent, which showed the market began in favor of this outcome just before the match opening.
But the trends of Feature 0 for $draw$ and $win$ were not agreed.

Now let us check what our model learned from such complicated curves.
\begin{figure}[hpbt]
\centering
\includegraphics[scale=0.5]{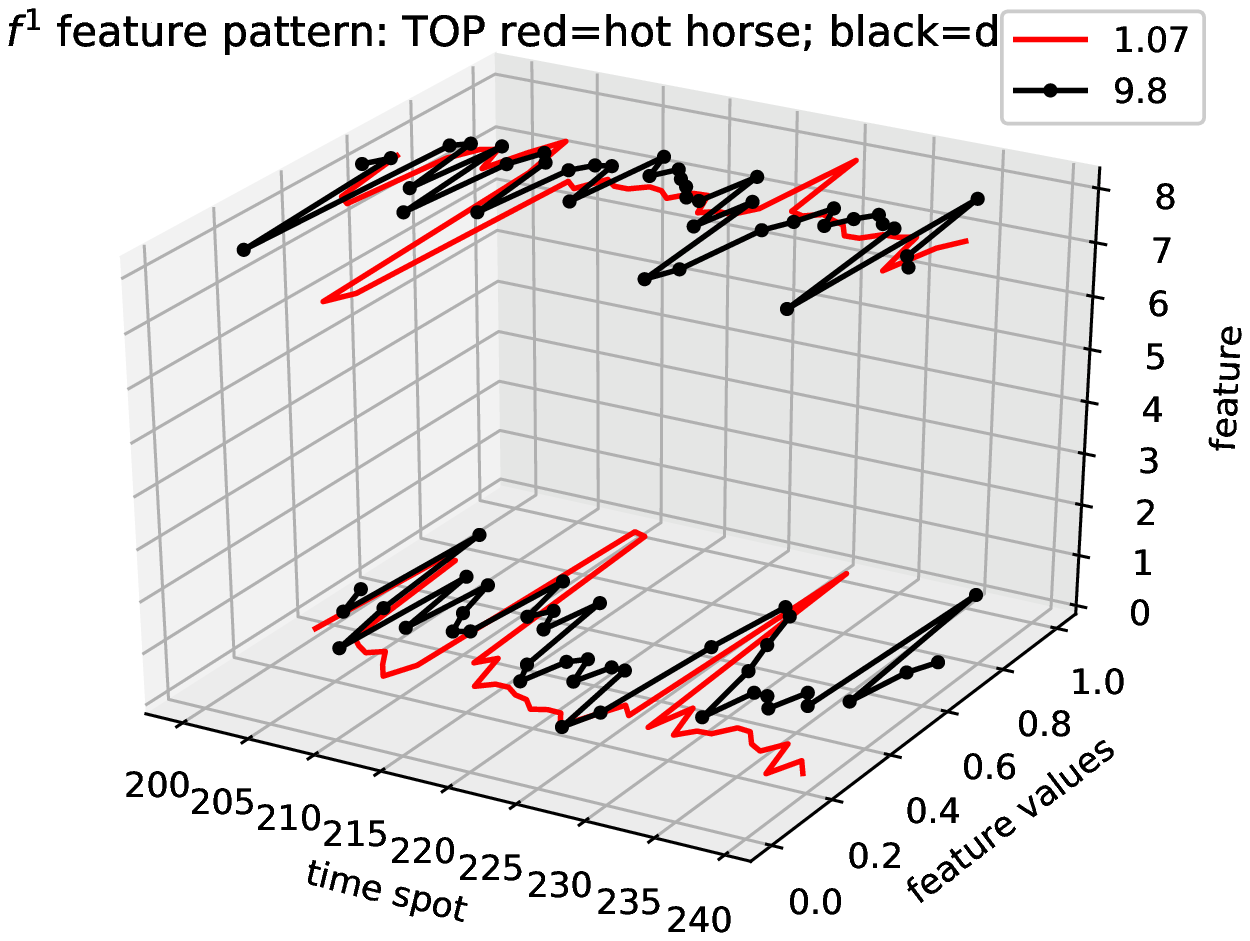}
\includegraphics[scale=0.5]{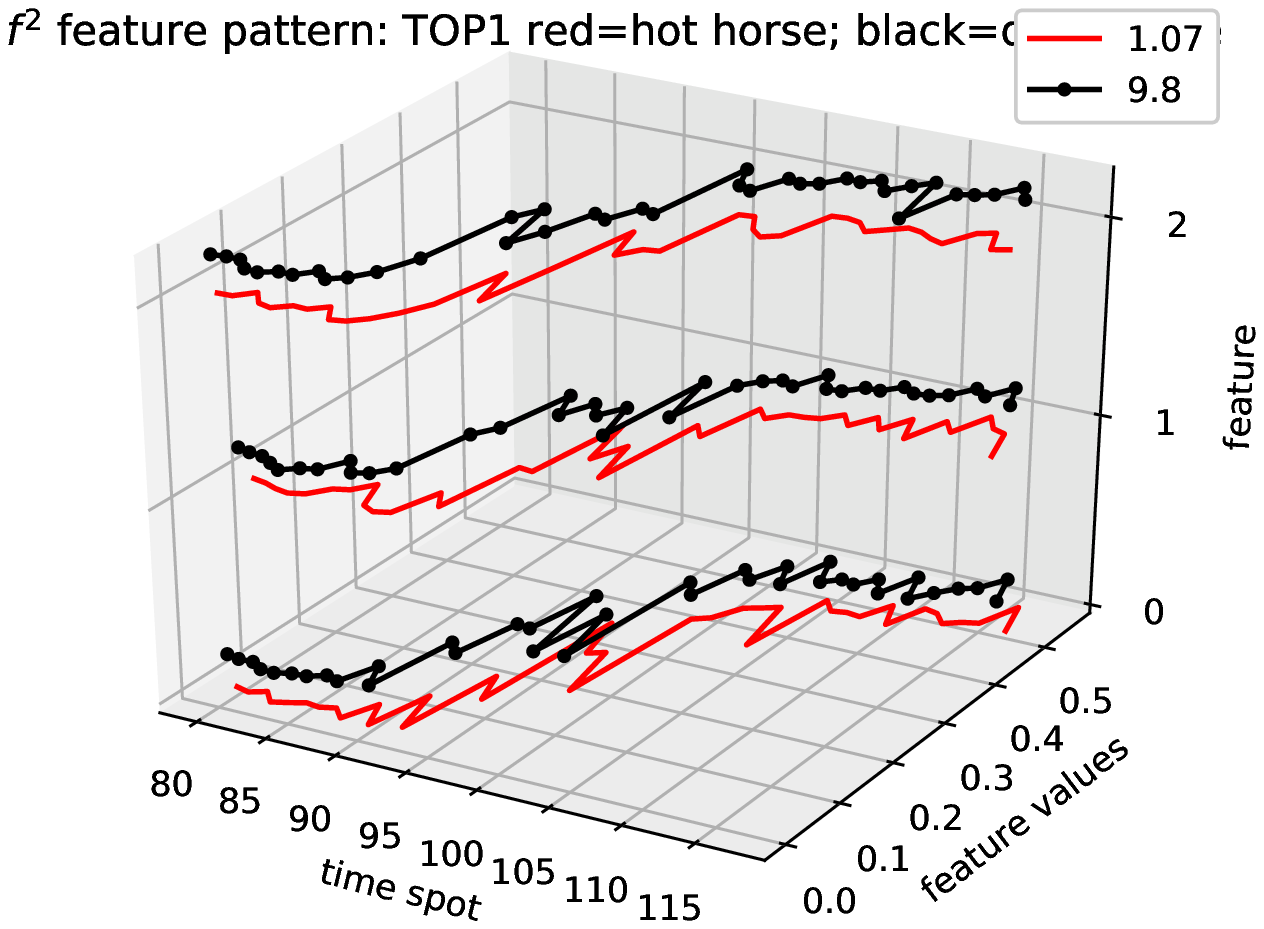}
\includegraphics[scale=0.5]{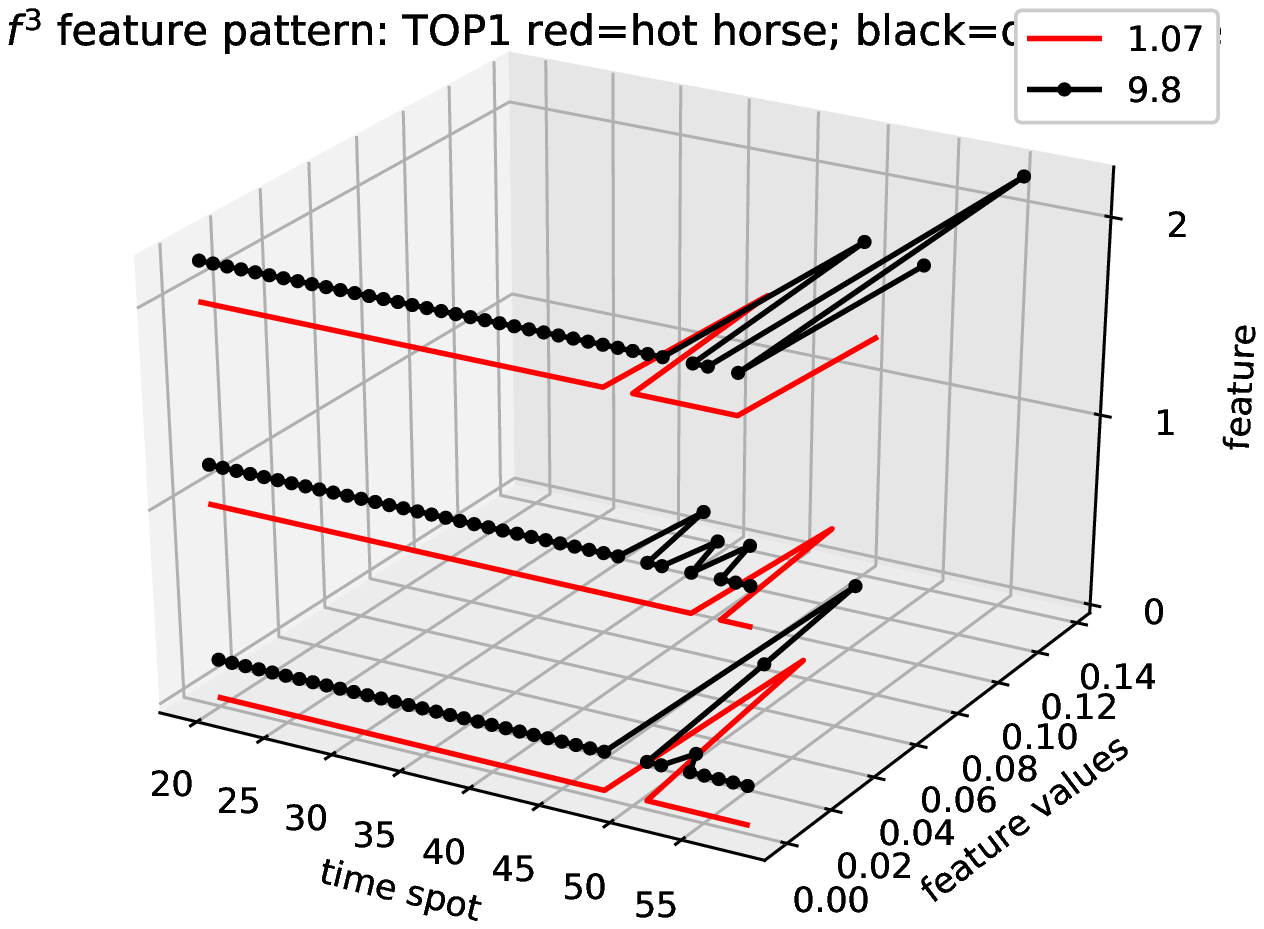}
\caption{CNN block features learned by our model.
The top subfigure is from the 1st CNN layer, The middle from the 2nd CNN layer, and  the bottom from the 3rd CNN layer.}
\label{fig:cnn}
\end{figure}
Figure \ref{fig:cnn} showed features learned from CNN block by our model.
For the sake of simplicity, we only showed the last 40 points in Figure \ref{fig:cnn},
and only Feature 0 and 8 in the top subfigure.

$\bm f^1$ in equation (\ref{eq:cnn1}) was designed to encode 60-dimensional input space into 9-dimensional space.
And such encoding was done by a non linear transformation in order to
map the sparse input space into dense feature space.
It made sense that each feature in $\bm f^1$ in Figure \ref{fig:cnn} was trying hard to re-represent the input feature.

$\bm f^2$ in equation (\ref{eq:locallyC}) was designed to capture the local features related to time spot.
We did not share the parameters of the convolution operand $\mathbb C$ along with the time axis.
Instead, we let each time spot have an independent convolution operand.
This gave each time spot a chance to embed different local patterns.
$\bm f^2$ showing in Figure \ref{fig:cnn} clearly illustrated that even the hottest horse had rich local feature patterns.
Comparing with the relatively stable curves in Figure \ref{fig:fea}, $\bm f^2$ features of the hottest horse were not so invariant as of the darkest horse.

$\bm f^3$ was designed to capture the global curve trend as the representative feature of the whole sequence.
It was consistent with human's regular sense that two patterns were revealed in the bottom subfigure of Figure \ref{fig:cnn}.
First, more far from the match beginning, more stable for the feature values.
Second, the feature curves of dark horse exhibited more fluctuation.

The extracted sequential $\bm f^3$ was feed to RNN block to mine the statable features showed in Figure \ref{fig:rnn}.
Finally, based on $\bm f^4$ feature pattern, MLP block gave the probability of 0.5777 to $draw$ outcome for that darkest horse in Table \ref{tab:pred},
while a not-so-low probability of .4075 to $win$ outcome and a neglect to $lose$.

\begin{figure}[hptb]
 \centering
  \includegraphics[scale=0.5]{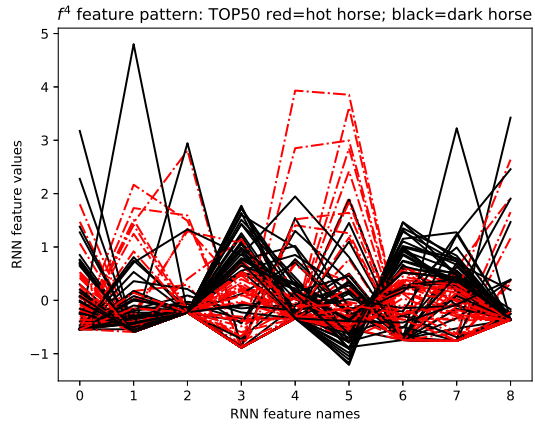}\\
  \caption{Feature patterns learned from RNN block for top 50 hottest and darkest horses}
  \label{fig:top50}
\end{figure}

We further gave feature patterns found by RNN block for top 50 hottest and darkest horses in Figure \ref{fig:top50}.
It seemed that patterns determined by Feature 3, 5, 6 and 7 showed differently between hot and dark horses.
Considering that these patterns would be further classified by MLP block,
it was acceptable that clearly distinguishable bound could not be found between these patterns.

\section{Discussion}
This paper built a learning model for detecting dark horse pattern from the sequential trading data.
We obtained good gains despite of prediction accuracy less than 50\%.
Further interesting insight was the possible patterns differentiated between dark and hot horses.
We developed an analytic framework to identify indicators of dark horses of soccer games.

The relationship from the input to output seemed to be conceptually correct in this study.
However, due to the existence of non linear transformations in three layers of our model,
the quantity relationship is hard to be identified clearly and simply.
In fact, deep learning model has always been criticized for its ``black box'' magic.
Since applying techniques of convolutional neural network to computer vision tasks has successfully resulted in interpretable visual features,
we are hoping that such successful possibility might exist in the domain of identifying dark horse.
How to map the mined features to semantic meanings for human to understand will always be challenge in machine learning domain.

Since the prediction market data flow is just like that on stock markets,
we hope our approach might be broadly applied to similar researches.

\bibliographystyle{unsrt}
\bibliography{goldfb2}

\end{document}